# Using Satellite Imagery for Good: Detecting Communities in Desert and Mapping Vaccination Activities

Anza Shakeel
Information Technology University
anzashakeel92@gmail.com

Mohsen Ali
Information Technology University
mohsen.ali@itu.edu.pk

## Abstract

*Deep convolutional neural networks (CNNs) have outperformed existing object recognition and detection algorithms. On the other hand satellite imagery captures scenes that are diverse. This paper describes a deep learning approach that analyzes a geo referenced satellite image and efficiently detects built structures in it. A Fully Convolution Network (FCN) is trained on low resolution Google earth satellite imagery in order to achieve end result. The detected built communities are then correlated with the vaccination activity that has furnished some useful statistics.*

## 1. Introduction

In today's era, geo-located maps fuel a considerable portion of the digital economy. From ride sharing services like Uber to finding restaurants in your neighborhood, we are dependent upon comprehensive, up to date and accurate maps. Aside from these commercial examples, accurate maps play a vital role in running many functions of the government. They are used for city planning, monitoring development and also in organizing rescue operations in times of disasters. The sheer usefulness and economic importance of these maps have pushed both government and non-government bodies to invest in creating, managing and updating these maps. Ideas like using mobile-phone-activity to geo-locate roads or directly crowdsourcing (Open Street Map, Wikimapia, etc..) the map information have been used to build and update maps. In the developing countries, government maintained maps are mostly not geo-located or are not readily accessible in the digitized format. This is true especially for remote regions of the developing countries, where mostly crowd-sourced information fails to be helpful. Satellite imagery provides opportunity to extract much of the information like urban and non-urban population, roads, crop yield without placing humans on ground. However, using humans to tag information by eyeballing the satellite imagery is a painstakingly difficult task.

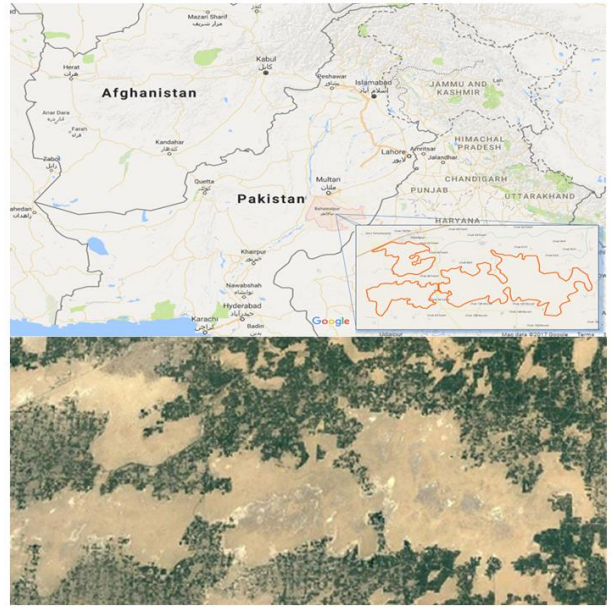

Figure 1: (Top) Polygon of the Cholistan region of Punjab is zoomed in and displayed on map view of Pakistan. (Bottom) Satellite view of the Cholistan region is shown.

Government of Punjab has been revamping vaccination program to increase its geographical coverage and to better manage vaccination staff spread across whole Punjab. To analyze the coverage of the vaccination activities especially in the areas far away from the urban centers of the Punjab, we need information about locations and size of communities in those regions. Since this information is not readily available, we have employed deep learning to detect house-like-structures on the freely available low resolution satellite imagery (visible spectrum) of the Cholistan Desert (Figure 1).

Our algorithm detects areas that are clustered together (to represent housing-clusters) and polygon fitting is performed so that we can create manageable representations (you can visualize our results on the interactive map below and technical details are discussed in the last section). Finally we map vaccination and public school location data over these detected regions. For vaccination data we calculate both count of vaccination activity per quarter per region, as well as coverage to understand how geographically diverse activities were performed in those region. Our analysis shows that although not all the detected regions appear to be covered, large numbers of them are. Some get vaccination activity in all the quarters; others are missed in one or two of the quarters. Large number of detected regions had at-least one public school as per government record. Certain detected regions were found to be getting very few or none of vaccination activity, their location and cluster indicates to us that there could be anomaly in the data; we hope to verify and correct that in our future version. We are in process of conducting such survey for whole Punjab, on much higher resolution satellite imagery.

## 2. Data-Set Details

In order to deeply investigate built-structure detection in satellite imagery, we had to be selective in terms of the data-set. Villages in desert areas are difficult to detect as the texture of roof-tops often merge with the ground, hence these regions demand very fine detection. Keeping this issue in mind we opted VillageFinders data-set for training [1]. The data is marked for segmentation purpose and includes nucleated villages from fifteen countries located mainly in Africa and Asia. The training data-set has a total of 300 images of size 512×512 with their respective ground-truths as binary images (pixels that represent a village area are 1 and others are marked 0).

To deal with our case of detection, we extracted multiple patches of size 128×128 from random locations of each image. These patches were then resized to 64×64 and were also jittered. During the process of patch extraction, patches from respective ground-truth binary images were also generated and a threshold of 0.75 was selected to decide whether the patch contained a building or not. Patches that had number of ones greater than 0.75 were assigned class label 1 (built) and patches that had number of ones less than 0.1 were assigned class label 0 (non-built). Sample patches from training set are shown in figure 2. The training data is normalized channel wise meaning that a random set of patches were selected from the training set and a mean image was computed from this set. The mean image was then subtracted from every training example as well as from each testing area.

We have used a Matlab® library that saves geo-located images from Google Earth™ when given maximum and minimum location of the area. This library is used to get images for Cholistan region of Punjab to test our trained model.

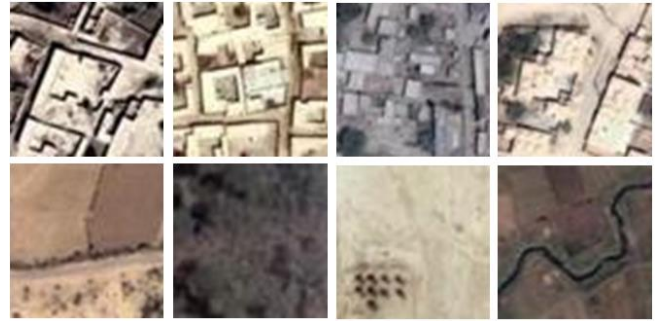

Figure 1: (Top) some of the patches from Class label 1 are shown. (Bottom) similarly patches from Class label 0.

## 3. Implementation Details

### 3.1. Network Architecture

Our initial step was to fine tune a pre-trained deep neural network and so we selected the VGG network [2]. After changing the last fully connected layer according to our classes, the error was only propagated till the fully connected layers. Training was turned off for all previous layers. The results from this trained model were not good this is due to the fact that VGG model is trained on imagenet data and our training examples are totally different. Due to this we decided to re-train the network on our data. The classification results were reasonable but as we had softmax in the end of fully connected layer the network was deciding on a patch.

Though we had two hundred thousand patches in the training set, still we had doubts that our results might improve more if we had more data as convolutional neural networks are known to be data hungry. In order to cater the need of more data, we augmented data during training. This was done by pausing training after every ten epochs and the model was tested with a portion of training data. All false negative examples were then added in the training set. This procedure was repeated five times.

In images taken from above, each pixel represents some area on ground. We wanted to preserve that information so we converted all three fully connected layers to convolutional layers [3]. In this way the network is now able to decide on each pixel. Instead of adding an up-sampling layer in the network, we have used bilinear interpolation to up-sample the probability maps to the size of input. For our final network (shown in figure 3) we have not used the batch normalization layer as it was distorting the results this was happening because satellite imagery has much more variance than a convolutional neural network can handle and so the batch normalization layers were not working. Also we first had Relu layer applied in the end of last conv layer which was later removed because it was making the output noisy. This we realized after visualizing the textures each layer was learning.

All model files are written in python using Chainer Framework. The network is trained on a 3 GB Nvidia gpu and took a week to get fully trained.

### 3.2. Post Processing Details

Once probability maps are interpolated they are mapped on their corresponding input images to see the detected regions. Initially the model is tested on both VillageFinder testing set and also on the area of Model Town Lahore images saved from Google Earth™. Results are shown in figure 4.

To define neighborhoods we fitted polygons on the results. This was done by applying threshold on the probability maps. Probabilities greater than or equal to 0.5 were said to be 1 and values less than threshold were marked zero. Methods of dilation and erosion were performed in order to clean the binaries before fitting polygons on them. As the testing images from Google Earth™ are geo referenced, we converted our results to world coordinates and wrote a kml of the polygons. Before polygon fitting we also tested our results and divided the world coordinates of our region of interest in small cells. These cells where given a value that was equal to the average probability of pixels covering a cell. While writing kml for this, the cell probability values where multiplied by the color of each cell. In this way we were able to visualize areas which the network detected with higher probability and also the regions which it detected with lower probability.

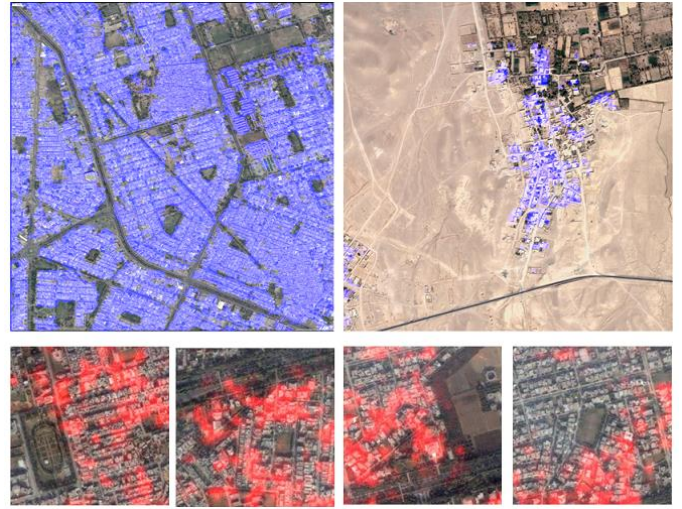

Figure 4: (Top-Left) Final Results on Model Town, Blue areas are detected built structures. (Top-Right) Final Results on VillageFinder testing image. (Bottom) Some initial results computed by fine-tuning only fully-connected layers of the VGG network, red areas are detected built structures.

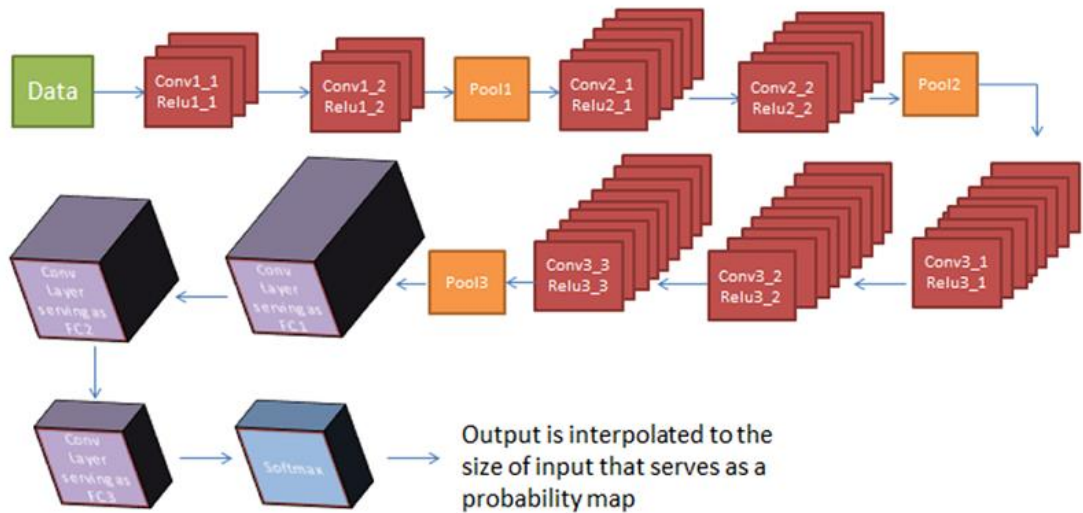

Figure 3: Network Architecture.

### 3.3. Testing Network on Cholistan Region

Cholistan region of Punjab Pakistan was selected to run testing experiments due to its difficult landscape. The exact location of Cholistan was noted from the Open Street Maps website and images were saved using Matlab®. The images were then passed from our trained model and then kml was generated as discussed in previous section (shown in figure 5).

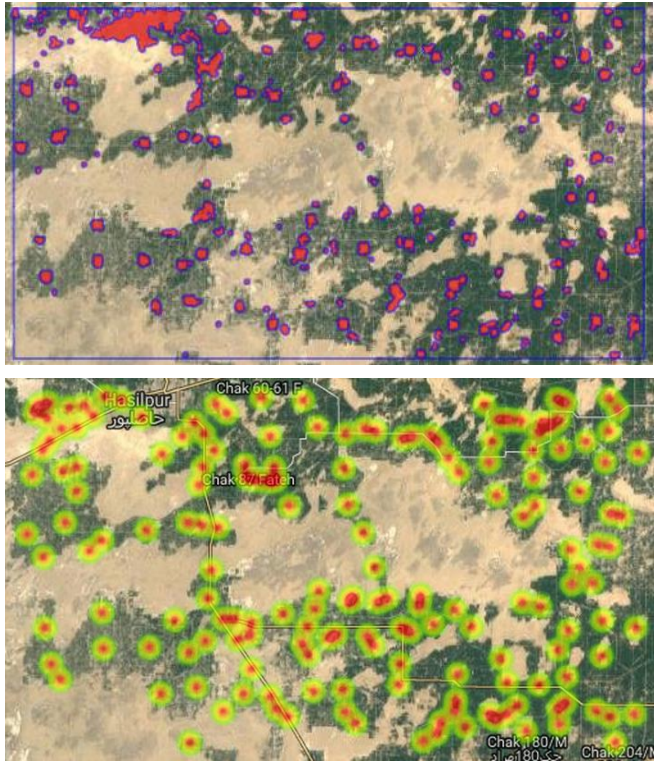

Figure 5: (Top) Detected segments of the Cholistan region of Punjab. Pink areas show detected regions. (Bottom) Heatmap of the test region.

### 3.4. Correlating Vaccination Activity

Pakistan has been aggressively running vaccination programs to decrease child mortality rate and eradicate diseases like polio. Government trained vaccinators travel to different parts of cities and far away villages to inoculate children in cities and far away villages. Punjab Information Technology Board (PITB) has distributed smartphones to these officials and a phone based application records information about the inoculation activity. Using this vaccination activity data, we analyze how often and in what ratio vaccinators visit different identified regions. We have only used data that is starting from August 2015 till December 2016, and have divided them into the interval of 3 months. Number of vaccinators who accessed a detected cluster is computed and a histogram is plotted to show the distribution of assigned vaccinators in the Cholistan region. Further a buffer of 200 meters is created around each location tagged by the vaccinator and then intersection of these buffers is computed resulting in an approximate movement of vaccinators. Percentage coverage is computed using movement of vaccinators in fixed area of detected segments.

As per Government's vaccination programme each vaccinator maintains a record that contains age of children with their pending vaccines. Following this record vaccinators visit their assigned vicinities to complete a vaccine's round according to its specificity. During each activity a histogram (shown in figure 7) is plotted showing number of vaccinators who accessed a detected community. These numbers justify the fact that regions with larger areas were accessed the most.

The movement of vaccinators covering an area in each community is proportioned over the area of detected region. These coverage percentages are graphed giving us segments that are covered most (shown in figure 8). A pattern of coverage can be seen based on the movement of vaccinators over the span.

### 3.5. Mapping Public Schools Data

Just like vaccination activity information, government schools data was correlated with the detected regions to understand accessibility to education. It was discovered that a large number of the detected regions did have the government school in them. Red areas show polygons with no schools, whereas the shades of green represent increasing number of schools with light green being the least (one or two) and dark green with most (more then 10) schools.

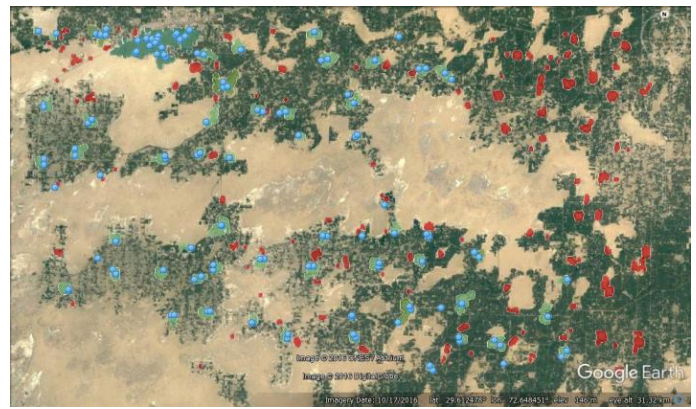

Figure 6: Strength of public schools mapped on detected segments.

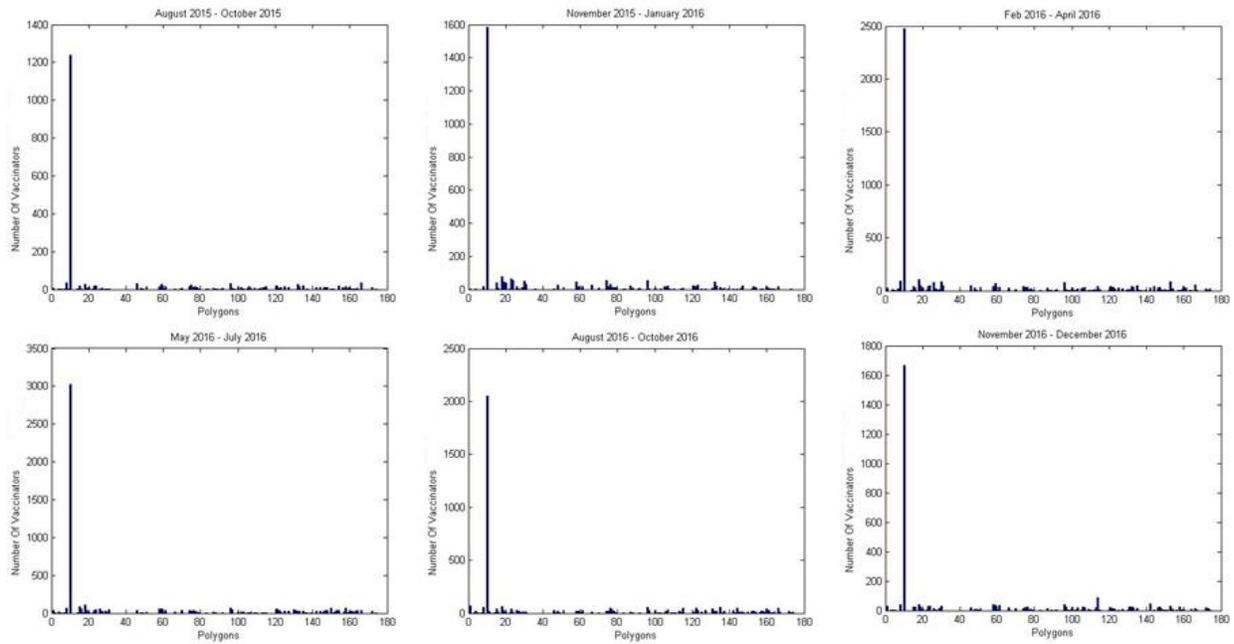

Figure 7: Histogram of vaccinators per segment

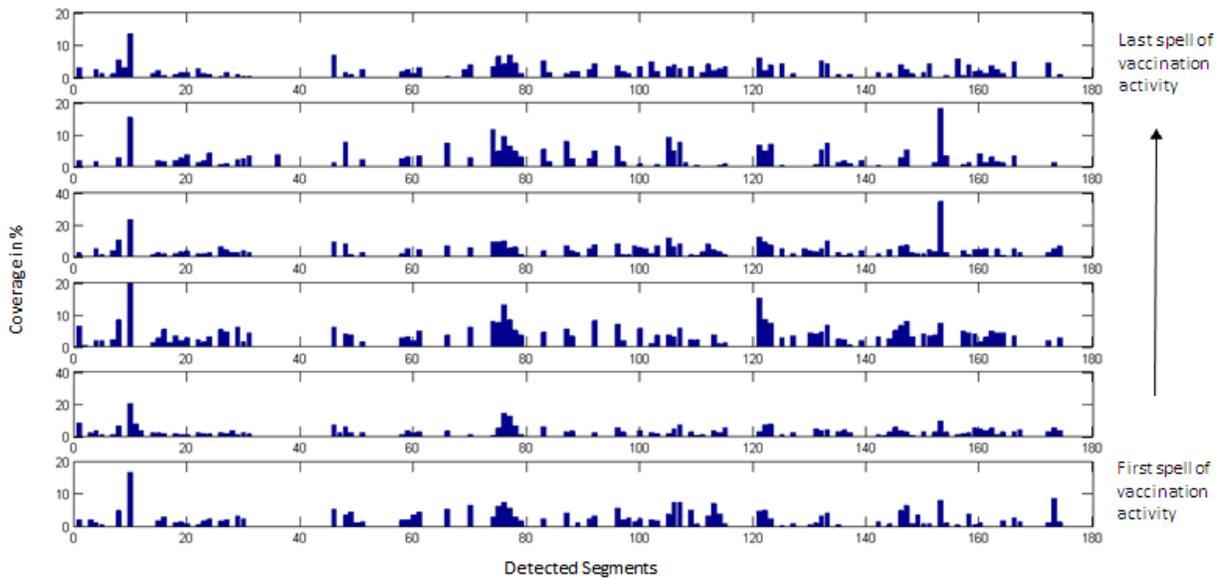

Figure 8: Histogram of area covered per detected segment during an activity (three month interval)

## 4. Future Work

Satellite imagery offers a vast research area. To take full advantage of the diversity these images contain, we plan to study the variety of bands that hyperspectral imagery presents. Detecting built structures successfully has inaugurated problems like house counting, tracking urbanization and detecting damaged structures. We intend to deeply analyze the detected segments in order to locate hospitals, schools and parks near each community.

## 5. Acknowledgements

We are grateful to Punjab Information Technology Board (PITB) especially Ms. Maria Badar for sharing vaccination activity data as well as public schools data.